# Application of the Modified Fractal Signature Method for Terrain Classification from Synthetic Aperture Radar Images

A. Malamou, C. Pandis, P. Frangos, P. Stefaneas, A. Karakasiliotis and D. Kodokostas
*National Technical University of Athens,*
*9, Iroon Polytechniou Str., 157 73 Zografou, Athens, Greece*
pfrangos@central.ntua.gr

*Abstract*— In this paper the Modified Fractal Signature (MFS) method is applied to real Synthetic Aperture Radar (SAR) images provided to our research group by SET 163 Working Group on SAR radar techniques. This method uses the 'blanket' technique to provide useful information for SAR image classification. It is based on the calculation of the volume of a 'blanket', corresponding to the image to be classified, and then on the calculation of the corresponding Fractal Area curve and Fractal Dimension curve of the image. The main idea concerning this proposed technique is the fact that different terrain types encountered in SAR images yield different values of Fractal Area curves and Fractal Dimension curves, upon which classification of different types of terrain is possible. As a result, a classification technique for five different terrain types (urban, suburban, rural, mountain and sea) is presented in this paper.

*Index Terms*— Classification of SAR images, Modified Fractal Signature (MFS) Method, Pattern Recognition, Synthetic Aperture Radar (SAR) Images.

## I. INTRODUCTION

Fractals describe infinitely complex patterns that are self-similar at different scales and are used as a mathematical tool for different applications, such as image analysis and classification, applied electromagnetism etc. [1]-[5]. The self - similar structure at many different scales is a basic characteristic of fractals. This characteristic may occur in either a statistical or an exact sense [2]. Therefore fractals can describe a high degree of geometrical complexity in several groups of data as well as in images. Images, and Synthetic Aperture Radar (SAR) images in particular, can be considered as fractals for a certain range of magnifications. Moreover, fractal objects have unique properties and characteristics that can be related to their geometric structure [1].

Hence, fractal analysis of SAR radar images in particular, which correspond to different terrain types, derived from real SAR radar data, can provide interesting classification and characterization results. For example, a SAR image of an urban area in comparison with a SAR image of a rural area, is expected to exhibit different properties, when they are treated as fractal objects.

In this paper the Modified Fractal Signature (MFS) method is applied to real spaceborne SAR radar images, provided to us by an International Working Group on SAR radar techniques (SET 163 Working Group). The main idea concerning this technique is the fact that different terrain types encountered in SAR images yield different characteristic values of 'Fractal Area' curves ($A_\delta$) and 'Fractal Dimension' (or 'Fractal Signature') curves ($F_D$) in particular, through which classification of different types of terrain is possible [3]-[9].

## II. MATHEMATICAL FORMULATION OF THE MFS METHOD

The Modified Fractal Signature (MFS) method is applied at images and it computes the values of 'Fractal Area' ($A_\delta$) and 'Fractal Dimension' (or 'Fractal Signature') ($F_D$) at different scales $\delta$ of the original image (hence a 'multi – resolution' approach [3]), by using an algorithm that incorporates the so called 'blanket' technique [3] – [5]. The images are initially converted to a gray – level function $g(x,y)$. In the 'blanket' approach all points of the three - dimensional space at distance $\delta$ or less from the gray level function $g(x,y)$ are considered. These points construct a 'blanket' of thickness $2\delta$ covering the initial gray level function. The covering blanket is defined by its upper surface $u_\delta(x,y)$ and its lower surface $b_\delta(x,y)$, as it is shown in Fig. 1 [4].

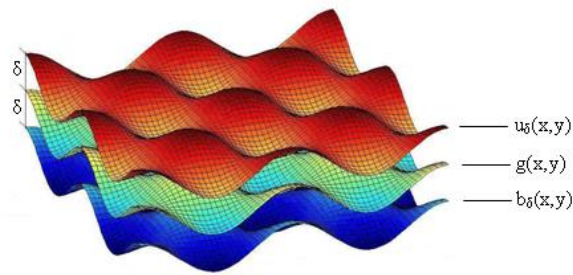

Fig. 1. 'Blanket' of thickness $2\delta$ defined by its upper $u_\delta(x,y)$ and lower $b_\delta(x,y)$ surface.

The upper and lower surface can be computed using an iterative algorithm ($\delta$ iterations). At first, the iteration number $\delta$ equals to zero ($\delta=0$) and the gray-level function

Manuscript received April XX, 20XX; accepted April XX, 20XX.
This research has been co-financed by the European Union (European Social Fund) and the Greek National Funds through the Operational Program "Education and Lifelong Learning" of the National Strategic Reference Framework (NSRF) – Research Funding Program: THALIS.

equals to the upper and lower surfaces, namely: $u_o(x,y)= b_o(x,y)=g(x,y)$. For iteration $\delta=1,2,\ldots$ the blanket surfaces are calculated through the following iterative formulae:

$$u_\delta(x,y) = \max\{u_{\delta-1}(x,y)+1, \max_{|(m,n)-(x,y)|\leq 1} u_{\delta-1}(m,n)\}$$
$$b_\delta(x,y) = \min\{b_{\delta-1}(x,y)-1, \min_{|(m,n)-(x,y)|\leq 1} b_{\delta-1}(m,n)\} \quad (1)$$

The image pixels $(m, n)$ with distance less than one from pixel $(x,y)$ are chosen in this paper as the four immediate neighbors of pixel $(x,y)$ [3]. Equation (1) ensures that the new upper surface $u_\delta$ is higher than $u_{\delta-1}$ by at least one. Likewise, the new lower surface $b_\delta$ is lower than $b_{\delta-1}$ by at least one [3].

Subsequently, the volume of the 'blanket' is calculated from $u_\delta(x,y)$ and $b_\delta(x,y)$ by:

$$Vol_\delta = \sum_{(x,y)} (u_\delta(x,y) - b_\delta(x,y)) \quad (2)$$

Furthermore, the 'Fractal Area' ($A_\delta$) can be calculated as following [3]-[5]:

$$A_\delta = \frac{Vol_\delta}{2\delta} \quad \text{or} \quad A_\delta = \frac{Vol_\delta - Vol_{\delta-1}}{2} \quad (3)$$

The 'Fractal Dimension' [or 'Fractal Signature' [3]] ($F_D$) can be calculated by the fractal area ($A_\delta$) using the following formula:

$$A_\delta \approx \beta \delta^{2-F_D} \quad (4)$$

where $\beta$ is a constant. In other words the 'Fractal Dimension' ($F_D$) corresponds to the rate of decreasing of the 'Fractal Area' ($A_\delta$) with increasing iteration $\delta$. Subsequently, from (4) it can be easily derived [4] that the 'Fractal Dimension' ($F_D$) can be obtained as a slope of the function $A_\delta$ in log-log scale, according to the formula:

$$F_D \approx 2 - \frac{\log_2 A_{\delta_1} - \log_2 A_{\delta_2}}{\log_2 \delta_1 - \log_2 \delta_2} \quad (5)$$

where in this paper we selected for convenience $\delta_1=1$ and $\delta_2=2,3,4\ldots$[3]-[5].

It appears that the value of 'Fractal Dimension' [or 'Fractal Signature'] ($F_D$) contains more information about the fractal properties of each terrain type than the value of 'Fractal Area' ($A_\delta$) [3] regarding the classification of different types of terrain in SAR images, and this is exactly the quantity which is used for image classification purposes [3] – [5].

### III. NUMERICAL RESULTS – TRAINING DATA

In this paper, the Modified Fractal Signature (MFS) method is applied for different types of terrain which are encountered in real field SAR images provided to us by an International Working Group on SAR techniques, named 'SET 163 Working Group'.

In Figs. 2-5 the SAR images used in the fractal analysis described above are shown. The images are real radar images (spaceborne SAR) related to four (4) different geographic regions in the United States of America (USA), namely in the city of New York, the city of Washington D.C., the city of Las Vegas and the state of Colorado.

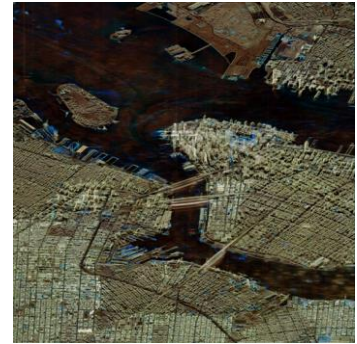

Fig.2. SAR image of the city of New York, USA.

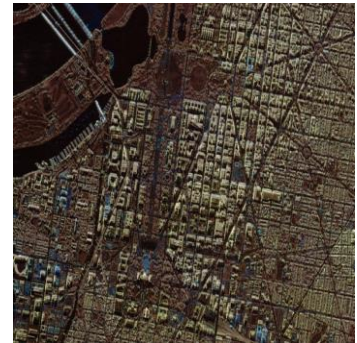

Fig.3. SAR image of the city of Washington D.C., USA.

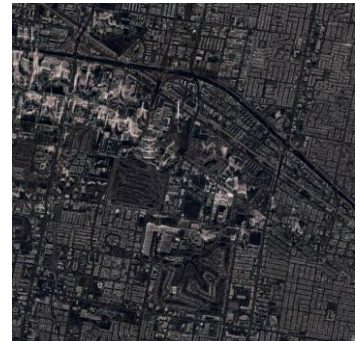

Fig.4. SAR image of the city of Las Vegas, USA.

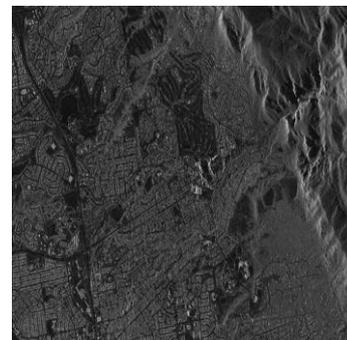

Fig.5. SAR image from a region at the state of Colorado, USA.

From the SAR images mentioned above, and in order that our proposed terrain classifier is constructed, twenty (20) sub-images of the same size were extracted. These twenty

(20) sub-images were organized in five (5) groups, each one of them corresponding to the five (5) different terrain types selected for this terrain classifier, namely for the following terrain types : urban site, suburban site, rural site, mountain site and sea site (i.e. 5 different types of terrain site). In other words, four (4) sub-images per terrain type were selected from the above mentioned SAR images, and the average of them was used for the construction of our classifier. All twenty (20) sub-images mentioned above actually represent the so – called 'training data' of our proposed classifier.

The 'Fractal Area' curves ($A_\delta$) for all twenty (20) sub - images of terrain mentioned above were calculated, and the average 'Fractal Area' curve for each type of terrain (out of 5) was calculated. After that, the corresponding 'multiresolution' curves for these five (5) types of terrain are shown in Fig. 6, in log – log scale (all the logarithms mentioned in this paper have as base the number two). Subsequently, through the use of (5), the corresponding 'Fractal Dimension' [or 'Fractal Signature' [3]] ($F_D$) curves were calculated, as shown in Fig. 7.

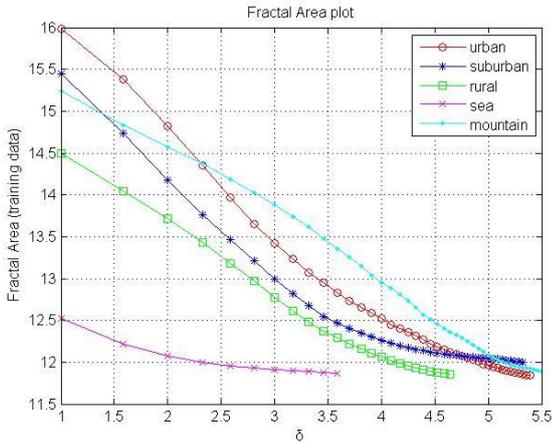

Fig. 6. 'Fractal Area' ($A_\delta$) versus iteration $\delta$ for each type of terrain (training data) in a log-log scale. Each curve is the average of four curves.

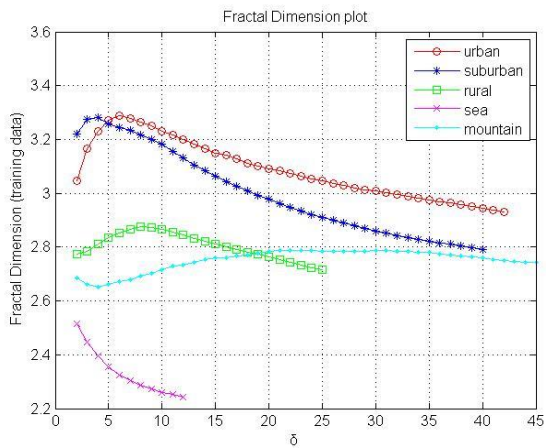

Fig. 7. 'Fractal Dimension' ($F_D$) versus iteration $\delta$ for each type of terrain (training data) obtained from 'Fractal Area' values, Fig. 6, by using eq. (5).

The curves in Fig. 7 show a clearly different pattern (with respect to 'Fractal Dimension' values and form of the corresponding curve) for each of the five (5) selected terrain types. As it will be explained in Section IV below, this will provide to us the basis for the construction of our terrain classifier, based on the 'distance' between 'Fractal Dimension' curves, in (6), below.

Moreover, the mean values and the standard deviation values for the gray-scale functions corresponding to the twenty (20) sub- images mentioned above are calculated. More precisely, the mean value and standard deviation value for each one of these twenty (20) sub-images was calculated, and subsequently the average value of mean and standard deviation values were calculated for each terrain type [out of the five (5) terrain types mentioned above]. These values are presented in Table I.

TABLE I. MEAN AND STANDARD DEVIATION VALUES.

|  | Mean Value | Standard Deviation Value |
|---|---|---|
| **urban (training data)** | 89.50 | 7.98 |
| **suburban (training data)** | 73.43 | 4.31 |
| **rural (training data)** | 44.55 | 3.50 |
| **mountain (training data)** | 60.51 | 15.82 |
| **sea (training data)** | 14.20 | 1.37 |

These differences in the mean and standard deviation values between the five (5) terrain types are reflected in a more detailed way in the 'Fractal Area' ($A_\delta$) curves and 'Fractal Dimension' ($F_D$) 'multi-resolution' curves presented in Fig. 6 and 7. As a result, the fractal analysis presented above can be used for the classification and the discrimination of different terrain types encountered in SAR images, as each it will be explained in Section IV below.

## IV. CLASSIFICATION RESULTS

For the classification purposes, five sub-images [of the same size with the 'training data' discussed in previous Section] were obtained from the same SAR images presented in Figs. 2-5, and these five sub – images represent here our 'testing data'. Each 'testing sub - image' corresponds to a particular terrain type, out of the five (5) discussed in the previous Section, namely : urban, suburban, rural, mountain and sea sites.

The 'testing data' sub - images were compared to the 'training data' sub - images based on their 'distance $D$' in the corresponding 'Fractal Dimension' curves ($F_D$). Namely, for two sub - images $i$ and $j$ with 'Fractal Dimension' curves $F_{Di}(\delta)$ and $F_{Dj}(\delta)$ respectively, the 'distance $D$' between them was computed using the following formula [3] :

$$D(i,j) = \sum_\delta \left[ \left(F_{Di}(\delta) - F_{Dj}(\delta)\right)^2 \cdot \log\left(\frac{\delta + \frac{1}{2}}{\delta - \frac{1}{2}}\right) \right] \quad (6)$$

where *δ* represents the number of iteration.

The above formula was applied to all possible pairs of sub images between the 'training data', Fig. 7, and the newly selected 'testing data'. The calculated 'distances *D*' for all such pairs are shown in Table II. A terrain type is identified by choosing the smallest 'distance *D*' from the corresponding 'training data'.

Then, from Table II we conclude that the same terrain types between 'training' and 'test' data exhibit the smallest 'distance *D*' in 'Fractal Dimension' curves ($F_D$), thus providing correct classification results in the classification experiment performed here. In other words, minimum value of 'distance *D*' were found on the diagonal of the 'classification matrix' ('confusion matrix') of Table II, thus providing correct classification results for the case examined here.

TABLE II. CLASSIFICATION MATRIX.

|  | urban (test data) | suburban (test data) | rural (test data) | mountain (test data) | sea (test data) |
|---|---|---|---|---|---|
| urban (training data) | 0.0164 | 0.0666 | 0.4318 | 0.7412 | 2.5293 |
| suburban (training data) | 0.1122 | 0.0149 | 0.4681 | 0.8797 | 2.6879 |
| rural (training data) | 0.6305 | 0.5277 | 0.0157 | 0.0297 | 0.8359 |
| mountain (training data) | 1.0736 | 0.9252 | 0.1304 | 0.0076 | 0.4991 |
| sea (training data) | 2.0747 | 2.0585 | 0.7091 | 0.3443 | 0.0016 |

## V. CONCLUSIONS

In this paper a novel approach for the classification of different terrain types which appear in SAR radar images of the terrain is described. This classification scheme is based on the calculation of 'Fractal Dimension' [i.e. 'Fractal Signature' [3]] 'multi – resolution' curves ($F_D$) for corresponding sub – images, and comparison of 'training' and 'testing' data (curves mentioned above) through calculation of the corresponding 'distance *D*' between them, in (6). Correct classification results were obtained for the classification experiment performed in this paper, based on real – life spaceborne SAR radar images.

As a future research in this area, more terrain data based on SAR radar images could be obtained for both 'training' and 'test' datasets, in order to build a more robust and more reliable terrain type classifier. Furthermore, more types of terrain structure could be introduced [than the five (5) types used in this paper], thus introducing a more sophisticated terrain classifier by using SAR radar data. Finally, other more advanced fractal methods than the 'MFS blanket' method presented in this paper, such as the 'Regny spectrum' method [1],[5] could be used for the purpose of terrain classification using SAR radar images.

## ACKNOWLEDGMENT

The authors (AM, AK, PF) would like to acknowledge SET 163 Working Group, and its Chairman Dr. Luc Vignaud (ONERA, France) in particular, for providing to us the real field Synthetic Aperture Radar (SAR) images shown in Figs. 2-5 above. In particular, these real SAR images were provided to SET 163 Working Group by the 'German Airospace Center, DLR' (spaceborne images). To the above institutes and involved scientists we express our sincere thanks for providing these real field radar images to us, in the framework of SET 163 Working Group.

Furthermore, the authors would like to express their sincere thanks to Prof. N. Ampilova and Prof. I. Soloviev, Faculty of Mathematics and Mechanics, St. Petersburg State University, Russia, for very interesting discussions and very helpful suggestions of theirs, which stimulated this research at the various stages of its implementation.